\documentclass[fleqn,10pt]{wlscirep}
\title{Atomic Convolutional Networks for Predicting Protein-Ligand Binding Affinity}

\author[1,+]{Joseph Gomes}
\author[2,+]{Bharath Ramsundar}
\author[3]{Evan N. Feinberg}
\author[1,*]{Vijay S. Pande}
\affil[1]{Department of Chemistry, Stanford University}
\affil[2]{Department of Computer Science, Stanford University}
\affil[3]{Program in Biophysics, Stanford University School of Medicine}
\affil[*]{pande@stanford.edu}
\affil[+]{these authors contributed equally to this work}

\usepackage{float}
\usepackage{amsmath}
\usepackage{rotating}

\begin{abstract}
Empirical scoring functions based on either molecular force fields or cheminformatics descriptors are widely used, in conjunction with molecular docking, during the early stages of drug discovery to predict potency and binding affinity of a drug-like molecule to a given target.  These models require expert-level knowledge of physical chemistry and biology to be encoded as hand-tuned parameters or features rather than allowing the underlying model to select features in a data-driven procedure.  Here, we develop a general 3-dimensional spatial convolution operation for learning atomic-level chemical interactions directly from atomic coordinates and demonstrate its application to structure-based bioactivity prediction.  The atomic convolutional neural network is trained to predict the experimentally determined binding affinity of a protein-ligand complex by direct calculation of the energy associated with the complex, protein, and ligand given the crystal structure of the binding pose. Non-covalent interactions present in the complex that are absent in the protein-ligand sub-structures are identified and the model learns the interaction strength associated with these features. We test our model by predicting the binding free energy of a subset of protein-ligand complexes found in the PDBBind dataset and compare with state-of-the-art cheminformatics and machine learning-based approaches.  We find that all methods achieve experimental accuracy (less than 1 kcal/mol mean absolute error) and that atomic convolutional networks either outperform or perform competitively with the cheminformatics based methods. Unlike all previous protein-ligand prediction systems, atomic convolutional networks are end-to-end and fully-differentiable.  They represent a new data-driven, physics-based deep learning model paradigm that offers a strong foundation for future improvements in structure-based bioactivity prediction. 
\end{abstract}
\begin{document}

\flushbottom
\maketitle

\thispagestyle{empty}
\section*{Introduction}

The success rate of the initial phases of drug discovery depends on the prediction, or measurement, of the affinity of a candidate ligand for a therapeutic target (e.g., protein) of interest. The space of synthetically accessible small molecules is unfathomably vast, estimated at over $10^{60}$ compounds \cite{virshup2013stochastic}. While exploring this entire space is currently computationally intractable, this combinatorial explosion underscores a core challenge in drug discovery: testing the affinity of as many small molecules as possible while maintaining a sufficient degree of accuracy. There is currently a significant trade-off in both experimental and computational drug screening approaches between speed, cost, and accuracy.\cite{dragiev2011systematic, wang2015accurate, trott2010autodock}

The introduction of machine learning into drug discovery pipelines has improved virtual drug screening as well as other physics-based evaluations of small molecules. Models such as random forests, logistic regression, and support vector machines \cite{svetnik2003random} have been used  extensively in virtual screening and chemoinformatics in the past decade and beyond. However, such models were beset by a fundamental limitation: the need to represent molecules with hand-curated, fixed-length vector features.

More recently, deep learning has demonstrated the potential to exceed this inherent limitation in traditional machine learning. The flexibility of deep neural networks allows models in principle to ``learn" successively higher orders of features from the simplest possible representations of the data at hand. In the computer vision field, for example, convolutional neural networks \cite{krizhevsky2012imagenet} applied to images can learn how to detect edges in early layers in the network; eyes, ears, etc. in intermediate network layers; and finally faces in terminal layers of the network. While such advanced artificial neural network frameworks have led to immense advances in the fields of computer vision and natural language processing, they have only recently penetrated other areas. 

The promise of DNNs has only just begun to be realized in the fields of chemistry and physics. Scientists have demonstrated the use of neural networks for molecular force fields \cite{behler2007nnpes,behler2011atomnn}, prediction of electronic properties of small molecules \cite{montavon2013machine}, protein-ligand binding \cite{durrant2011nnscore, wallach2015atomnet,ragoza2016protein}, chemoinformatics \cite{duvenaud2015convolutional, kearnes2016molecular, altae2016low}, among others. Previous works that addressed protein-ligand binding have used simpler, fully connected neural networks based on hand-curated features \cite{durrant2011nnscore} or used a direct application of voxel-based convolutional neural networks in the two-class classification task of distinguishing small molecule binders from non-binders. \cite{wallach2015atomnet,ragoza2016protein}  The atomic convolution architecture was inspired by the atomic fingerprint neural network, originally proposed by Behler and Parrinello \cite{behler2007nnpes,behler2011atomnn} for the purpose of fitting potential energy surfaces, with several key differences.  Rather than choose fixed parameters for the featurization of molecules into atomic fingerprint vectors, we optimize these parameters simultaneously with the neural network in an end-to-end fashion to allow the model to make data-driven decisions on features that are important to predicting ligand binding.  In addition, we utilize a neighbor list routine to reduce the computational cost of model training and evaluation, allowing the models to scale well to large systems.  This work represents the largest application of this method to date and the first application based purely on experimental data.

In this paper, we introduce a new fundamental deep learning primitive to improve the representational and predictive power for molecular systems, the Atomic Convolutional Neural Network (ACNN). Analogous to the original CNNs \cite{krizhevsky2012imagenet, lecun1995comparison}, ACNN's directly exploit the local three-dimensional structure of molecules to hierarchically learn more complex chemical features by optimizing both the model and featurization simultaneously in an end-to-end fashion.
Whereas many previous works focus on two-class classification of drug binding or non-binding, we apply ACNNs to directly predict binding free energy arising from non-covalent interactions purely from experimental data by direct integration of a ligand-binding thermodynamic cycle into the model optimization.  The resulting optimized energy function is size-extensive and fully-differentiable with respect to atomic coordinates.  In addition, ACNN models have the desirable feature of being able to generalize to larger systems than those it has been trained on.  After describing the mathematical architecture of the ACNNs, we report their application for the prediction of protein-ligand binding affinity with the publicly available PDBBind dataset. \cite{wang2005pdbbind}  All calculations were done in the open-source molecular machine learning package DeepChem.\cite{deepchem}  We open source all datasets and code required to reproduce this work.

\section*{Methods}

\subsection*{Atomic Convolutional Neural Networks}
\begin{figure}
    \centering
    \includegraphics[width=\textwidth]{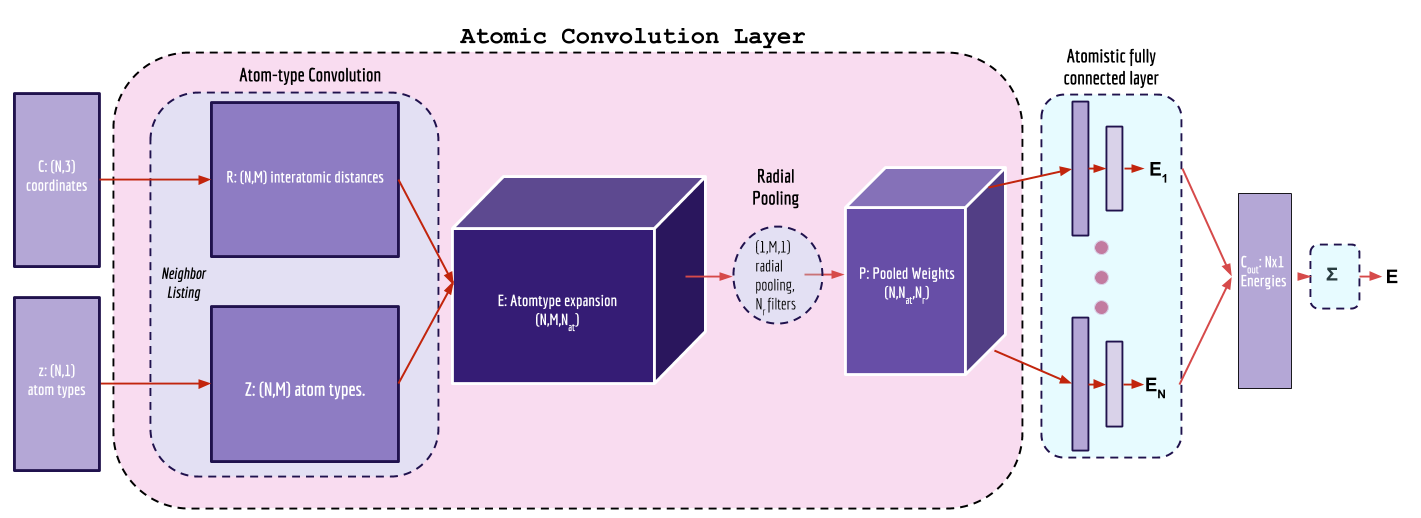}
    \caption{Schematic of atomic convolution layer.}
    \label{fig:atomic_conv}
\end{figure}

The architecture of the atomic convolutional neural network is shown in Figure~\ref{fig:atomic_conv}.  We introduce two new primitive convolutional operations, atom type convolution and radial pooling. The atom type convolution makes use of a neighbor-listed distance matrix to extract features encoding local chemical environments from an input representation (Cartesian atomic coordinates) that does not necessarily contain spatial locality.

\subsubsection*{Distance matrix and neighbor list construction}
The distance matrix $R$ and atomic number matrix $Z$ are constructed from the Cartesian atomic coordinates $X$.  We use a neighbor list routine \cite{yip1989neighbor} to reduce the complexity of this step from $O(N^2)$ to $O(NM)$, where $M$ is some maximum number of neighbors.  The radial interaction cutoff used in the neighbor list routine is 12 Å and the neighbor list is truncated at number of neighbors $M$, typically 12.  The initial neighbor list distance matrix representation is invariant to rigid body translation and rotation of the molecular system but not atom index or neighbor list atom index permutation.

The distance matrix construction accepts as input a $(N,3)$ coordinate matrix $C$. This matrix is ``neighbor listed'' into a $(N,M)$ matrix $R$. For atom $i$, the neighbor listing process finds the $M$ atoms closest to $i$ spatially. Let $N_i = [a_{i_1},\dotsc,a_{i_m}]$ be the list of neighbors. Then $R_{i,j}$ is defined as
\begin{eqnarray*}
R_{i,j} &= \|C_i - C_{i_j}\|_2
\end{eqnarray*}
The neighbor listing operation also constructs from the $(N,1)$ atomic number vector $z$ a $(N,M)$ matrix $Z$ which lists the atomic number of neighboring atoms (atom types).
\begin{equation*}
Z_{i,j} = \textrm{Atom type of }a_{i_j}
\end{equation*}

\subsubsection*{Atom type convolution} 
The output of the atom type convolution is constructed from the distance matrix $R$ and atomic number matrix $Z$. The matrix $R$ is fed into a (1x1) filter with stride 1 and depth of $N_\textrm{at}$, where $N_\textrm{at}$ is the number of unique atomic numbers (atom types) present in the molecular system.  The atom type convolution kernel is a step function that operates on neighbor distance matrix $R$:
\begin{equation*}
(K*R)_{i, j} = R_{i,j}K_{i,j}^a
\end{equation*}
where 
\begin{equation*}
K_{i,j}^a = \begin{cases} 
      1 & Z_{i,j} = N_a \\
      0 & \textrm{otherwise}
   \end{cases}
\end{equation*}
where $N_a$ is the atomic number of atom type $a$ ($a = 1,\dotsc,N_\textrm{at}$).  The atom type convolution layer is applied to the neighbor distance matrix to create output matrix $E$ of shape $(N, M, N_\textrm{at})$. The atom type convolution can also be thought of as an expansion layer that one-hot encodes the atom type $N_{a}$ into separate copies of the distance matrix $R_{N_{a}}$.
\subsubsection*{Radial pooling layer}
Radial pooling is a dimensionality reduction process intended to down-sample the output of the atom type convolution. This dimensionality reduction is done in part to prevent over-fitting by providing an abstracted form of the representation through feature binning, as well as reducing the number of parameters learned.  In addition, radial pooling provides an output representation invariant to neighbor list atom index permutation.  

Radial pooling takes input tensor $E$ output by the atom-type convolution of shape $(N,M,N_{at})$. Radial pooling is then performed by applying a radial filter over non-overlapping sub-regions of the input representation.  Mathematically, radial pooling layers pool over tensor slices (receptive fields) of size (1x$M$x1) with stride $1$ and a depth of $N_r$, where $N_r$ is the number of desired radial filters.  Radial pooling filters are of the functional form $f_s$

\begin{eqnarray*}
f_s(r_{i,j}) &= \exp\left(-\frac{(r_{i,j} - r_s)^2}{\sigma_s^2}f_c(r_{i,j})\right)\\
f_c(r_{i,j}) &= \begin{cases} 
      \frac{1}{2}\cos\left (\frac{\pi r_{i,j}}{R_c}\right) & 0 < r_{i,j} < R_c \\
      0 & r_{i,j} \geq R_c
   \end{cases}
\end{eqnarray*}

Parameters $r_s$ and $\sigma_s$ are learn-able parameters for pooling function $f_s$. Parameter $R_c$ is the radial interaction cutoff, which is fixed to 12 Å.  Then, the output pooled matrix $P$ is of shape $(N,N_\textrm{at},N_r)$ and has entries
\begin{eqnarray*}
P_{i,n_a,n_r} = \beta_{n_r} \sum_{j=1}^M f_{n_r}(E_{i,j,n_a}) + b_{n_r}
\end{eqnarray*}
where $\beta_{n_r}$ is the non-learn-able scaling constant and $b_{n_r}$ is a non-learn-able bias constant.  Conceptually, applying radial pooling following an atom type convolution layer produces features which sum the pairwise-interactions between atom $i$ with atom type $a_i$ (e.g. H, C, N, etc.) and all neighboring atoms of type $a_j$ (e.g. H-H, H-C, H-N, etc.). 

\subsubsection*{Atomistic fully connected network}
The output $P$ of the radial pooling layer has shape $(N,N_\textrm{at}, N_r)$. We flatten the coordinates for each atom to obtain a tensor of shape $(N, N_\textrm{at}\cdot N_r)$. Atomic convolution layers can be stacked by feeding the flattened output of the radial pooling layer back into the atom-type convolution operation. Finally, we feed the tensor row-wise (per-atom) into a fully-connected network. The same fully connected weights and biases are used for each atom in a given molecule. The output of the atomistic fully-connected network is energy $E_i$ of atom $i$ ($i=1\cdots N$).  The total energy of the molecule, $E = \sum_i E_i$, is the sum of the atomic energies and is consequently invariant to atom index permutation.  Since the fully connected network input dimension only depends on the number of features, not number of atoms, a fully trained ACNN model can generalize to systems larger than those contained in the training set, given a fixed number of atom types and radial pooling filters.

\subsection*{Application of atomic convolution networks to predicting protein-ligand binding affinity}
\begin{figure}[h]
  \centering
  \includegraphics[width=.8\textwidth]{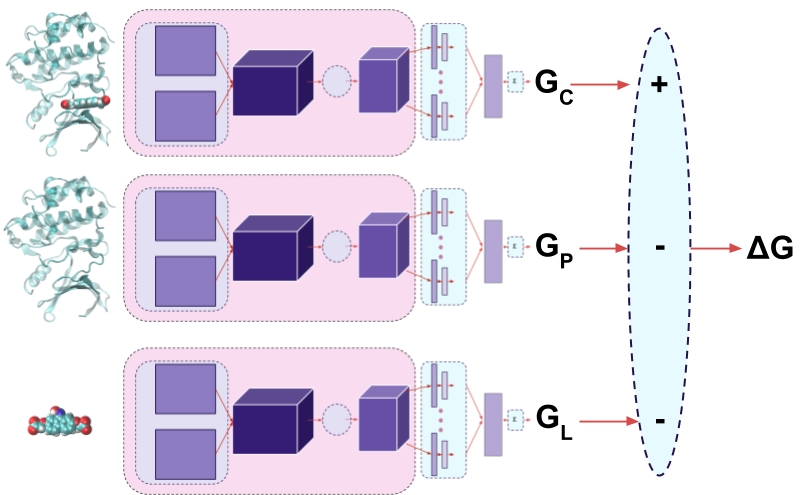}
  \caption{Diagram of Atomic Convolutions on Protein Ligand Systems.}
  \label{fig:protein_ligand}
\end{figure}
The atomic convolution network architecture produces an energy that is size-extensive and differentiable with respect to atomic positions. In order to learn non-covalent interactions using the atomic convolution energy function, we integrate the following thermodynamic cycle into the learning process
\begin{eqnarray*}
\Delta G_\textrm{complex} = G_\textrm{complex} - G_\textrm{protein} - G_\textrm{ligand}
\end{eqnarray*}

To create a model which can accurately predict $\Delta G_{\textrm{complex}}$, we create three weight-sharing, replica networks, one each for complex, protein, and ligand (Figure~\ref{fig:protein_ligand}). The full system is trained against the loss
\[\mathcal{L} = (\Delta G_\textrm{complex} - y_\textrm{complex})^2\]
Note how the complete network incorporates the thermodynamic cycle as a subcomponent. This system is trained end-to-end to predict $\Delta G$, but does so in a fashion which respects the underlying adsorption thermodynamics.  ACNN models were trained with stochastic gradient descent with a batch size of 24 and the ADAM optimizer \cite{kingma2014adam} for 100 epochs.

\subsection*{Baseline Comparison}
\subsubsection*{Grid Featurizer Method}
The grid featurization method (GRID), first introduced in molecular machine learning benchmark MoleculeNet \cite{wu2017moleculenet}, was chosen as the baseline method for comparison of structure-based machine learning methods. The GRID method incorporates both the ligand and target protein structural information by considering both the chemical interaction within the binding pocket as well as features of the protein and ligand individually. The grid featurizer was inspired by the NNscore featurizer \cite{durrant2011nnscore} and SPLIF \cite{da2014splif} but optimized for speed, robustness, and generalizability. The molecular interactions enumerated by the GRID method include salt bridges and hydrogen bonding between protein and ligand, intra-ligand circular fingerprints\cite{rogers2010extended}, intra-protein circular fingerprints, and protein-ligand SPLIF fingerprints.

The features produced by GRID were used to create a random forest regression model (GRID-RF) and a fully-connected neural network model (GRID-NN). Random forest models were trained using the scikit-learn \texttt{RandomForestRegressor}.  Fully-connected neural network models were trained using the DeepChem \texttt{TensorflowMultitaskRegressor}. Model hyperparameters are described elsewhere \cite{wu2017moleculenet}.  The GRID method in conjunction with the random forest regression model is currently the top performing machine learning method for structure-based binding affinity prediction on the PDBBind dataset contained within the MoleculeNet benchmark.

\subsubsection*{Graph Convolution Method}

The graph convolutional neural network (GCNN) of Duvenaud and Maclaurin\cite{duvenaud2015convolutional} was chosen as a baseline for ligand-based convolutional deep learning methods.  Graph convolutional models treat molecules as undirected graphs whose vertices and edges represent individual atoms and bonds. A graph convolutional layer applies the same learnable function to every atom in the graph, resulting in a learned featurization. The graph convolutional algorithm is similar in spirit to the ACNN, but uses the bonded connectivity graph instead of a spatial neighbor list. Put another way, the GCNN is purely two dimensional, while the ACNN is natively a three dimensional architecture. Descriptions of the graph convolutional models and model hyperparameters are given elsewhere in the literature \cite{wu2017moleculenet}.

\subsubsection*{ECFP Fingerprint Method}

The extended-connectivity fingerprints\cite{rogers2010extended} (ECFP) featurization method was chosen as a baseline for ligand-based machine learning methods.  ECFPs are widely-used molecular characterizations in chemical informatics. The ECFP featurization process decomposes a  molecule into segments originated from non-hydrogen atoms, each assigned with a unique identifier. These segments and identifiers are extended through bonds to generate larger substructures and corresponding identifiers. The substructures are hashed into a fixed length binary fingerprint.

The features produced by ECFP were used to create a random forest regression model (ECFP-RF) and a fully-connected neural network model (ECFP-NN). Random forest models were trained using the scikit-learn \texttt{RandomForestRegressor}.  Fully-connected neural network models were trained using the DeepChem \texttt{TensorflowMultitaskRegressor}. Model hyperparameters are described elsewhere \cite{wu2017moleculenet}.

\subsubsection*{Datasets}
PDBBind \cite{wang2004pdbbind,wang2005pdbbind} is a database of experimentally measured binding affinities for protein ligand complexes. PDBBind provides 3D crystal structures and associated inhibition constant $K_i$ for the protein-ligand complexes in its collection. The PDBBind 2015 dataset contains three subsets: core (195 structures), refined (3,706 structures), and full (14,260). The crystal structures present in the refined datasets are obtained at a higher resolution and cleaned more thoroughly than the full dataset, in addition to more stringent requirements on the quality of the complex structure, quality of the binding data, and the nature of the complex. The complexes present in the refined set are then clustered by 90\% similarity in protein sequence to create 65 families.  From these families, the core set was created by selecting three representative complexes (weakest binding, median binding, and strongest binding) to control sample redundancy.  Additional information on the curation of these datasets can be found elsewhere \cite{liu2014pdb}.  We use the core and refined subsets to train and benchmark the performance of ACNNs.  Other popular protein-ligand scoring functions trained on the PDBBind dataset include X-Score\cite{wang2002further} and AutoDock Vina\cite{trott2010autodock}.

\subsubsection*{Dataset Splitting}
We consider four methods of splitting PDBBind core and refined sets into subsets for train/test evaluation: random, stratified, scaffold, and temporal. All train/test splits follow the 80/20 ratio.  Splits were performed with a fixed random seed for reproducibility and to ensure all models are trained and evaluated on the same data.  The random split randomly splits samples into train/test subsets.  The stratified split sorts examples in order of increasing inhibition constant $K_i$, and then choses samples 10 at a time and randomly splits these samples into train/test subsets to ensure that each set contains the full range of inhibition constant present in the parent dataset.  Scaffold splitting was performed using Bemis-Murcko ligand scaffold clustering \cite{bemis1996properties}.  This procedure clusters ligand molecules present in the datasets by removing side chain atoms and sorting by scaffold frequency.  Common scaffolds are placed in the train set and uncommon scaffolds are placed in the test set.  Since this split attempts to separate structurally distinct molecules into train and test sets, it represents a much greater challenge for learning algorithms compared with random and stratified split.  Temporal splitting was performed based on the year that the protein-ligand complex was entered in the Protein Data Bank.  This split tests the ability of the learning algorithm to use prior historical data to predict results of future experiments, similar to typical use in prospective drug discovery.  The random, stratified, and scaffold splits are explained in much greater detail in the MoleculeNet\cite{wu2017moleculenet} benchmark.

\subsubsection*{Performance Metrics}
To determine the train and test set performance, the squared Pearson correlation coefficient ($R^2$) of log$K_i$ and binding free energy mean unsigned error were evaluated.  Raw scatter plots comparing the actual and predicted log$K_i$ for all datasets and splits using the ACNN and GRID-RF methods are given in Supplementary Information.

\begin{table}[h]
  \centering
    \begin{tabular}{|c||c|c||c|c||c|c||c|c||c|c||c|c|}
    \hline
    \multicolumn{1}{|c||}{} & \multicolumn{2}{c||}{ACNN} & \multicolumn{2}{c||}{GRID-RF} & \multicolumn{2}{c||}{GRID-NN} & \multicolumn{2}{c||}{GCNN} & \multicolumn{2}{c||}{ECFP-RF} & \multicolumn{2}{c|}{ECFP-NN}\\
    \hline
    \hline
    Split & Train & Test & Train & Test & Train & Test & Train & Test & Train & Test & Train & Test \\
    \hline
    Random & .912 & \textbf{.448} & .969 & .336 & .963 & .058 & .676 & .265 & .920 & .212 & .942 & .227 \\
    Stratified & .939 & .116 & .969 & \textbf{.132} & .963 & .165 & .735 & .064 & .924 & .071 & .942 & .077 \\
    Scaffold & .911 & .043 & .965 & .109 & .953 & .067 & .797 & \textbf{.254} & .920 & .218 & .940 & .206 \\
    Temporal & .923 & .251 & .972 & \textbf{.287} & .957 & .245 & .744 & .095 & .925 & .206 & .952 & .071 \\
    \hline
    \end{tabular}
  \caption{Performance (Pearson $R^2$) on PDBBind core train/test sets.}
  \label{tab:core-pearson}
\end{table}
\begin{table}[h]
  \centering
    \begin{tabular}{|c||c|c||c|c||c|c||c|c||c|c||c|c|}
    \hline
    \multicolumn{1}{|c||}{} & \multicolumn{2}{c||}{ACNN} & \multicolumn{2}{c||}{GRID-RF} & \multicolumn{2}{c||}{GRID-NN} & \multicolumn{2}{c||}{GCNN} & \multicolumn{2}{c||}{ECFP-RF} & \multicolumn{2}{c|}{ECFP-NN}\\
    \hline
    \hline
    Split & Train & Test & Train & Test & Train & Test & Train & Test & Train & Test & Train & Test \\
    \hline
    Random & 0.325 & 0.774 & 0.385 & \textbf{0.741} & 0.230 & 0.877 & 0.656 & 1.112 & 0.399 & 1.112 & 0.234 & 1.138 \\
    Stratified & 0.282 & 0.997 & 0.339 & 0.990 & 0.205 & \textbf{0.813} & 0.556 & 0.995 & 0.410 & 0.901 & 0.223 & 1.115 \\
    Scaffold & 0.410 & 0.993 & 0.338 & 1.397 & 0.211 & 1.630 & 0.516 & \textbf{0.883} & 0.438 & 1.003 & 0.221 & 0.909 \\
    Temporal & 0.363 & 0.974 & 0.368 & \textbf{0.860} & 0.237 & 0.809 & 0.588 & 1.062 & 0.413 & 0.974 & 0.341 & 1.265 \\
    \hline
    \end{tabular}
  \caption{Performance (MUE [kcal/mol]) on PDBBind core train/test sets.}
  \label{tab:core-mae}
\end{table}

\begin{table}[h]
  \centering
    \begin{tabular}{|c||c|c||c|c||c|c||c|c||c|c||c|c|}
    \hline
    \multicolumn{1}{|c||}{} & \multicolumn{2}{c||}{ACNN} & \multicolumn{2}{c||}{GRID-RF} & \multicolumn{2}{c||}{GRID-NN} & \multicolumn{2}{c||}{GCNN} & \multicolumn{2}{c||}{ECFP-RF} & \multicolumn{2}{c|}{ECFP-NN}\\
    \hline
    \hline
    Split & Train & Test & Train & Test & Train & Test & Train & Test & Train & Test & Train & Test \\
    \hline
    Random & .705 & .508 & .962 & \textbf{.546} & .976 & .539 & .581 & .403 & .883 & .466 & .850 & .386 \\
    Stratified & .793 & .491 & .963 & \textbf{.562} & .983 & .494 & .635 & .410 & .883 & .477 & .840 & .462 \\
    Scaffold & .752 & .267 & .964 & \textbf{.349} & .977 & .334 & .652 & .135 & .861 & .229 & .824 & .146 \\
    Temporal & .704 & \textbf{.529} & .963 & .486 & .978 & .455 & .596 & .315 & .875 & .401 & .827 & .362 \\
    \hline
    \end{tabular}
  \caption{Performance (Pearson $R^2$) on PDBBind refined train/test sets.}
  \label{tab:refined-pearson}
\end{table}
\begin{table}[h]
  \centering
    \begin{tabular}{|c||c|c||c|c||c|c||c|c||c|c||c|c|}
    \hline
    \multicolumn{1}{|c||}{} & \multicolumn{2}{c||}{ACNN} & \multicolumn{2}{c||}{GRID-RF} & \multicolumn{2}{c||}{GRID-NN} & \multicolumn{2}{c||}{GCNN} & \multicolumn{2}{c||}{ECFP-RF} & \multicolumn{2}{c|}{ECFP-NN}\\
    \hline
    \hline
    Split & Train & Test & Train & Test & Train & Test & Train & Test & Train & Test & Train & Test \\
    \hline
    Random & 0.518 & 0.653 & 0.240 & \textbf{0.630} & 0.468 & 1.208 & 0.598 & 0.726 & 0.319 & 0.661 & 0.335 & 0.740 \\
    Stratified & 0.540 &  0.686 & 0.237 & \textbf{0.650} & 0.439 & 1.109 & 0.559 & 0.710 & 0.319 & 0.660 & 0.344 & 0.678 \\
    Scaffold & 0.555 & 0.579 & 0.245 & \textbf{0.537} & 0.406 & 1.177 & 0.557 & 0.805 & 0.339 & 0.749 & 0.373 & 0.856 \\
    Temporal & 0.526 & 0.668 & 0.235 & \textbf{0.666} & 0.410 & 1.184 & 0.610 & 0.865 & 0.321 & 0.770 & 0.371 & 0.798 \\
    \hline
    \end{tabular}	
  \caption{Performance (MUE [kcal/mol]) on PDBBind refined train/test sets.}
  \label{tab:refined-mae}
\end{table}

\section*{Discussion and Results}
The results of model training and evaluation on the PDBBind core set are shown in Tables \ref{tab:core-pearson} and \ref{tab:core-mae}.  All ACNN core set models were trained with 15 atom types, 3 radial filters, and a (32, 32, 16) pyramidal atomistic fully-connected layer. ACNN models show comparable or better performance with GRID-RF across all data splits in both Pearson $R^2$ and mean absolute error metrics.  Mean absolute error metrics across all splits within the structure-based methods (ACNN, GRID-RF, GRID-NN) are all comparable, with the exception of scaffold split where ACNN models demonstrate better test performance.  The ligand-based baselines (GCNN, ECFP-RF, ECFP-NN) all exhibit strong train performance, but inferior test performance and fail to generalize well compared with structure-based models due to the lack of features representing protein structure.  One notable exception is the performance of GCNN on the scaffold split core test set, which performs best out of all methods across both metrics.

In all ACNN core set models, the mean absolute error on the test set is less than 1 kcal/mol, a barrier which has been previously cited as chemical accuracy necessary to enable rapid computational drug design.\cite{peterson2012chemical} The demonstrated performance indicates that ACNN models are able to learn chemical interactions from relatively few (less than 160) datapoints. However, in all core set models, there is a large gap between train and test performance, indicating that overfitting is a general problem in these models, and that either more data or better parameter regularization is required.

The results of model training and evaluation on the PDBBind refined set are shown in Tables \ref{tab:refined-pearson} and \ref{tab:refined-mae}.  All ACNN refined set models were trained with 25 atom types, 5 radial filters, and a (128, 128, 64) pyramidal atomistic fully-connected layer.  Learnable parameter regularization was enforced by the Dropout method \cite{srivastava2014dropout}, with a dropout probability of 40\% applied to all but the final fully connected layer.  ACNN models exhibit comparable Pearson $R^2$ and nearly identical mean absolute error performance on the refined datasets compared with GRID models.  The addition of Dropout to the ACNN models does improve generalizability between train and test sets across all splits, and the maximum mean unsigned error of the ACNN models across all test sets is less than 1 kcal/mol.  In general, end-to-end trained models (ACNN and GCNN) exhibit the best generalization between train and test. Ligand-based baselines do well compared to structure-based mathods on the refined dataset, also exhibiting mean unsigned error less than 1 kcal/mol across all splits.  Even without protein structure, these methods are able to discern what is and what isn't a potent drug-like molecule.    

The largest shortcoming of the ACNN models in the current work is the inability to learn better from larger PDBBind datasets. Since the ACNN models are featurized based on raw atomic coordinates, rather than interaction counts and molecular graph fingerprints, they are more susceptible to incorrect learning due to noise in experimental crystal structure.  ACNN models trained on the full PDBBind set, which has much looser requirements on the completeness and accuracy of structural data and binding constants, do not exhibit smooth loss convergence, and train/test performance can show large fluctuations even at long train times.  The degradation of performance due to structural data can be improved by including more, higher-quality crystal structures and binding free energy measurements or through systematic structural refinement with, for example, a molecular force field in addition to the inclusion of multiple ligand binding conformations per complex.

We anticipate the current ACNN models will exhibit worse performance when trained models are applied well-outside the scope of training data, forcing the model to extrapolate rather than interpolate.  This includes new atom types, interaction types, ligand scaffolds, and protein binding pockets that differ greatly than those included in the training set.  The poor performance on predicting the binding affinity of unseen scaffolds is an issue with all methods tested in this study.  The improved performance of ACNN models over traditional ML approaches on the core dataset suggest that ACNN models can be retrained with relatively few data points on systems of interest outside of the original training set to boost predictive power. 

Early experiments with weight regularization techniques such as Dropout and L2/L1 regularization suggest that model overfitting and test set performance are highly sensitive to regularization strength and technique.  Increasing number of parameters beyond what was presented in the paper leads to overfitting on train and reduced performance on test.  The addition of regularization techniques may also help ACNN models become less sensitive to noise in structural and thermodynamic data and reduce overfitting.  A systematic cross-validation study of model architecture and weight regularization is currently underway.

We argue that ACNNs represent a fundamental advance in protein-ligand affinity prediction due to the fact that they utilize a fully-differentiable end-to-end learned representation. In machine translation, Bahdanau et al.\cite{bahdanau2014neural} introduced the first end-to-end neural machine translation architecture. This architecture was the first fully-differentiable system for machine translation. In its first iteration, the model only performed at par with older machine translation systems (much as ACNNs only perform at par with simpler machine learning baselines). However, with two years of refinement, the latest neural machine translation system has been deployed in Google Translate\cite{wu2016google} and led to dramatic improvements of 60\% relative to the older system. We believe that a similar potential for improvement exists with atomic convolutions.

Finally, we note that ACNNs should also be applicable to a number of applications not directly related to drug discovery. For example, ACNNs may be able to improve upon existing neural network-based potentials for the automated fitting of potential energy surfaces (force fields), which have wide applicability in theoretical chemistry for accelerating \textit{ab-initio} molecular dynamics simulations of many-body systems.  By introducing a learnable, convolutional 3-D atomic fingerprint and neighbor list routine to the existing atomic neural network potential, the accuracy and computational tractability of models are vastly improved. Additionally, ACNNs are a general featurization technique that should perform well on traditional virtual screening tasks where 3-D structural information is required, such as the prediction of electronic properties of molecules \cite{montavon2013machine} and the virtual screening of new materials, such as organic photovoltaics \cite{hachmann2011harvard} and light-emitting diodes \cite{gomez2016design}.

\bibliography{sample}

\begin{thebibliography}{10}
\expandafter\ifx\csname url\endcsname\relax
  \def\url#1{\texttt{#1}}\fi
\expandafter\ifx\csname urlprefix\endcsname\relax\def\urlprefix{URL }\fi
\expandafter\ifx\csname doiprefix\endcsname\relax\def\doiprefix{DOI }\fi
\providecommand{\bibinfo}[2]{#2}
\providecommand{\eprint}[2][]{\url{#2}}

\bibitem{virshup2013stochastic}
\bibinfo{author}{Virshup, A.~M.}, \bibinfo{author}{Contreras-Garc{\'\i}a, J.},
  \bibinfo{author}{Wipf, P.}, \bibinfo{author}{Yang, W.} \&
  \bibinfo{author}{Beratan, D.~N.}
\newblock \bibinfo{title}{Stochastic voyages into uncharted chemical space
  produce a representative library of all possible drug-like compounds}.
\newblock \emph{\bibinfo{journal}{Journal of the American Chemical Society}}
  \textbf{\bibinfo{volume}{135}}, \bibinfo{pages}{7296--7303}
  (\bibinfo{year}{2013}).

\bibitem{dragiev2011systematic}
\bibinfo{author}{Dragiev, P.}, \bibinfo{author}{Nadon, R.} \&
  \bibinfo{author}{Makarenkov, V.}
\newblock \bibinfo{title}{Systematic error detection in experimental
  high-throughput screening}.
\newblock \emph{\bibinfo{journal}{BMC bioinformatics}}
  \textbf{\bibinfo{volume}{12}}, \bibinfo{pages}{25} (\bibinfo{year}{2011}).

\bibitem{wang2015accurate}
\bibinfo{author}{Wang, L.} \emph{et~al.}
\newblock \bibinfo{title}{Accurate and reliable prediction of relative ligand
  binding potency in prospective drug discovery by way of a modern free-energy
  calculation protocol and force field}.
\newblock \emph{\bibinfo{journal}{Journal of the American Chemical Society}}
  \textbf{\bibinfo{volume}{137}}, \bibinfo{pages}{2695--2703}
  (\bibinfo{year}{2015}).

\bibitem{trott2010autodock}
\bibinfo{author}{Trott, O.} \& \bibinfo{author}{Olson, A.~J.}
\newblock \bibinfo{title}{Autodock vina: improving the speed and accuracy of
  docking with a new scoring function, efficient optimization, and
  multithreading}.
\newblock \emph{\bibinfo{journal}{Journal of computational chemistry}}
  \textbf{\bibinfo{volume}{31}}, \bibinfo{pages}{455--461}
  (\bibinfo{year}{2010}).

\bibitem{svetnik2003random}
\bibinfo{author}{Svetnik, V.} \emph{et~al.}
\newblock \bibinfo{title}{Random forest: a classification and regression tool
  for compound classification and qsar modeling}.
\newblock \emph{\bibinfo{journal}{Journal of chemical information and computer
  sciences}} \textbf{\bibinfo{volume}{43}}, \bibinfo{pages}{1947--1958}
  (\bibinfo{year}{2003}).

\bibitem{krizhevsky2012imagenet}
\bibinfo{author}{Krizhevsky, A.}, \bibinfo{author}{Sutskever, I.} \&
  \bibinfo{author}{Hinton, G.~E.}
\newblock \bibinfo{title}{Imagenet classification with deep convolutional
  neural networks}.
\newblock In \emph{\bibinfo{booktitle}{Advances in neural information
  processing systems}}, \bibinfo{pages}{1097--1105} (\bibinfo{year}{2012}).

\bibitem{behler2007nnpes}
\bibinfo{author}{Behler, J.} \& \bibinfo{author}{Parrinello, M.}
\newblock \bibinfo{title}{Generalized neural-network representation of
  high-dimensional potential-energy surfaces}.
\newblock \emph{\bibinfo{journal}{Phys. Rev. Lett.}}
  \textbf{\bibinfo{volume}{98}}, \bibinfo{pages}{146401}
  (\bibinfo{year}{2007}).

\bibitem{behler2011atomnn}
\bibinfo{author}{Behler, J.}
\newblock \bibinfo{title}{Atom-centered symmetry functions for constructing
  high-dimensional neural network potentials}.
\newblock \emph{\bibinfo{journal}{The Journal of Chemical Physics}}
  \textbf{\bibinfo{volume}{134}}, \bibinfo{pages}{074106}
  (\bibinfo{year}{2011}).

\bibitem{montavon2013machine}
\bibinfo{author}{Montavon, G.} \emph{et~al.}
\newblock \bibinfo{title}{Machine learning of molecular electronic properties
  in chemical compound space}.
\newblock \emph{\bibinfo{journal}{New Journal of Physics}}
  \textbf{\bibinfo{volume}{15}}, \bibinfo{pages}{095003}
  (\bibinfo{year}{2013}).

\bibitem{durrant2011nnscore}
\bibinfo{author}{Durrant, J.~D.} \& \bibinfo{author}{McCammon, J.~A.}
\newblock \bibinfo{title}{Nnscore 2.0: a neural-network receptor--ligand
  scoring function}.
\newblock \emph{\bibinfo{journal}{Journal of chemical information and
  modeling}} \textbf{\bibinfo{volume}{51}}, \bibinfo{pages}{2897--2903}
  (\bibinfo{year}{2011}).

\bibitem{wallach2015atomnet}
\bibinfo{author}{Wallach, I.}, \bibinfo{author}{Dzamba, M.} \&
  \bibinfo{author}{Heifets, A.}
\newblock \bibinfo{title}{Atomnet: a deep convolutional neural network for
  bioactivity prediction in structure-based drug discovery}.
\newblock \emph{\bibinfo{journal}{arXiv preprint arXiv:1510.02855}}
  (\bibinfo{year}{2015}).

\bibitem{ragoza2016protein}
\bibinfo{author}{Ragoza, M.}, \bibinfo{author}{Hochuli, J.},
  \bibinfo{author}{Idrobo, E.}, \bibinfo{author}{Sunseri, J.} \&
  \bibinfo{author}{Koes, D.~R.}
\newblock \bibinfo{title}{Protein-ligand scoring with convolutional neural
  networks}.
\newblock \emph{\bibinfo{journal}{arXiv preprint arXiv:1612.02751}}
  (\bibinfo{year}{2016}).

\bibitem{duvenaud2015convolutional}
\bibinfo{author}{Duvenaud, D.~K.} \emph{et~al.}
\newblock \bibinfo{title}{Convolutional networks on graphs for learning
  molecular fingerprints}.
\newblock In \emph{\bibinfo{booktitle}{Advances in neural information
  processing systems}}, \bibinfo{pages}{2224--2232} (\bibinfo{year}{2015}).

\bibitem{kearnes2016molecular}
\bibinfo{author}{Kearnes, S.}, \bibinfo{author}{McCloskey, K.},
  \bibinfo{author}{Berndl, M.}, \bibinfo{author}{Pande, V.} \&
  \bibinfo{author}{Riley, P.}
\newblock \bibinfo{title}{Molecular graph convolutions: moving beyond
  fingerprints}.
\newblock \emph{\bibinfo{journal}{Journal of computer-aided molecular design}}
  \textbf{\bibinfo{volume}{30}}, \bibinfo{pages}{595--608}
  (\bibinfo{year}{2016}).

\bibitem{altae2016low}
\bibinfo{author}{Altae-Tran, H.}, \bibinfo{author}{Ramsundar, B.},
  \bibinfo{author}{Pappu, A.~S.} \& \bibinfo{author}{Pande, V.}
\newblock \bibinfo{title}{Low data drug discovery with one-shot learning}.
\newblock \emph{\bibinfo{journal}{arXiv preprint arXiv:1611.03199}}
  (\bibinfo{year}{2016}).

\bibitem{lecun1995comparison}
\bibinfo{author}{LeCun, Y.} \emph{et~al.}
\newblock \bibinfo{title}{Comparison of learning algorithms for handwritten
  digit recognition}.
\newblock In \emph{\bibinfo{booktitle}{International conference on artificial
  neural networks}}, vol.~\bibinfo{volume}{60}, \bibinfo{pages}{53--60}
  (\bibinfo{organization}{Perth, Australia}, \bibinfo{year}{1995}).

\bibitem{wang2005pdbbind}
\bibinfo{author}{Wang, R.}, \bibinfo{author}{Fang, X.}, \bibinfo{author}{Lu,
  Y.}, \bibinfo{author}{Yang, C.-Y.} \& \bibinfo{author}{Wang, S.}
\newblock \bibinfo{title}{The pdbbind database: methodologies and updates}.
\newblock \emph{\bibinfo{journal}{Journal of medicinal chemistry}}
  \textbf{\bibinfo{volume}{48}}, \bibinfo{pages}{4111--4119}
  (\bibinfo{year}{2005}).

\bibitem{deepchem}
\bibinfo{title}{Deepchem: Deep-learning models for drug discovery and quantum
  chemistry}.
\newblock \bibinfo{howpublished}{\url{https://github.com/deepchem/deepchem}}.
\newblock \bibinfo{note}{Accessed: 2017-03-28}.

\bibitem{yip1989neighbor}
\bibinfo{author}{Yip, V.} \& \bibinfo{author}{Elber, R.}
\newblock \bibinfo{title}{Calculations of a list of neighbors in molecular
  dynamics simulations}.
\newblock \emph{\bibinfo{journal}{Journal of Computational Chemistry}}
  \textbf{\bibinfo{volume}{10}}, \bibinfo{pages}{921--927}
  (\bibinfo{year}{1989}).

\bibitem{kingma2014adam}
\bibinfo{author}{Kingma, D.} \& \bibinfo{author}{Ba, J.}
\newblock \bibinfo{title}{Adam: A method for stochastic optimization}.
\newblock \emph{\bibinfo{journal}{arXiv preprint arXiv:1412.6980}}
  (\bibinfo{year}{2014}).

\bibitem{wu2017moleculenet}
\bibinfo{author}{Wu, Z.} \emph{et~al.}
\newblock \bibinfo{title}{Moleculenet: A benchmark for molecular machine
  learning}.
\newblock \emph{\bibinfo{journal}{arXiv preprint arXiv:1703.00564}}
  (\bibinfo{year}{2017}).

\bibitem{da2014splif}
\bibinfo{author}{Da, C.} \& \bibinfo{author}{Kireev, D.}
\newblock \bibinfo{title}{Structural protein–ligand interaction fingerprints
  (splif) for structure-based virtual screening: Method and benchmark study}.
\newblock \emph{\bibinfo{journal}{Journal of Chemical Information and
  Modeling}} \textbf{\bibinfo{volume}{54}}, \bibinfo{pages}{2555--2561}
  (\bibinfo{year}{2014}).

\bibitem{rogers2010extended}
\bibinfo{author}{Rogers, D.} \& \bibinfo{author}{Hahn, M.}
\newblock \bibinfo{title}{Extended-connectivity fingerprints}.
\newblock \emph{\bibinfo{journal}{Journal of chemical information and
  modeling}} \textbf{\bibinfo{volume}{50}}, \bibinfo{pages}{742--754}
  (\bibinfo{year}{2010}).

\bibitem{wang2004pdbbind}
\bibinfo{author}{Wang, R.}, \bibinfo{author}{Fang, X.}, \bibinfo{author}{Lu,
  Y.} \& \bibinfo{author}{Wang, S.}
\newblock \bibinfo{title}{The pdbbind database: Collection of binding
  affinities for protein- ligand complexes with known three-dimensional
  structures}.
\newblock \emph{\bibinfo{journal}{Journal of medicinal chemistry}}
  \textbf{\bibinfo{volume}{47}}, \bibinfo{pages}{2977--2980}
  (\bibinfo{year}{2004}).

\bibitem{liu2014pdb}
\bibinfo{author}{Liu, Z.} \emph{et~al.}
\newblock \bibinfo{title}{Pdb-wide collection of binding data: current status
  of the pdbbind database}.
\newblock \emph{\bibinfo{journal}{Bioinformatics}} \bibinfo{pages}{btu626}
  (\bibinfo{year}{2014}).

\bibitem{wang2002further}
\bibinfo{author}{Wang, R.}, \bibinfo{author}{Lai, L.} \& \bibinfo{author}{Wang,
  S.}
\newblock \bibinfo{title}{Further development and validation of empirical
  scoring functions for structure-based binding affinity prediction}.
\newblock \emph{\bibinfo{journal}{Journal of computer-aided molecular design}}
  \textbf{\bibinfo{volume}{16}}, \bibinfo{pages}{11--26}
  (\bibinfo{year}{2002}).

\bibitem{bemis1996properties}
\bibinfo{author}{Bemis, G.~W.} \& \bibinfo{author}{Murcko, M.~A.}
\newblock \bibinfo{title}{The properties of known drugs. 1. molecular
  frameworks}.
\newblock \emph{\bibinfo{journal}{Journal of medicinal chemistry}}
  \textbf{\bibinfo{volume}{39}}, \bibinfo{pages}{2887--2893}
  (\bibinfo{year}{1996}).

\bibitem{peterson2012chemical}
\bibinfo{author}{Peterson, K.~A.}, \bibinfo{author}{Feller, D.} \&
  \bibinfo{author}{Dixon, D.~A.}
\newblock \bibinfo{title}{Chemical accuracy in ab initio thermochemistry and
  spectroscopy: current strategies and future challenges}.
\newblock \emph{\bibinfo{journal}{Theoretical Chemistry Accounts}}
  \textbf{\bibinfo{volume}{131}}, \bibinfo{pages}{1079} (\bibinfo{year}{2012}).

\bibitem{srivastava2014dropout}
\bibinfo{author}{Srivastava, N.}, \bibinfo{author}{Hinton, G.~E.},
  \bibinfo{author}{Krizhevsky, A.}, \bibinfo{author}{Sutskever, I.} \&
  \bibinfo{author}{Salakhutdinov, R.}
\newblock \bibinfo{title}{Dropout: a simple way to prevent neural networks from
  overfitting.}
\newblock \emph{\bibinfo{journal}{Journal of Machine Learning Research}}
  \textbf{\bibinfo{volume}{15}}, \bibinfo{pages}{1929--1958}
  (\bibinfo{year}{2014}).

\bibitem{bahdanau2014neural}
\bibinfo{author}{Bahdanau, D.}, \bibinfo{author}{Cho, K.} \&
  \bibinfo{author}{Bengio, Y.}
\newblock \bibinfo{title}{Neural machine translation by jointly learning to
  align and translate}.
\newblock \emph{\bibinfo{journal}{arXiv preprint arXiv:1409.0473}}
  (\bibinfo{year}{2014}).

\bibitem{wu2016google}
\bibinfo{author}{Wu, Y.} \emph{et~al.}
\newblock \bibinfo{title}{Google's neural machine translation system: Bridging
  the gap between human and machine translation}.
\newblock \emph{\bibinfo{journal}{arXiv preprint arXiv:1609.08144}}
  (\bibinfo{year}{2016}).

\bibitem{hachmann2011harvard}
\bibinfo{author}{Hachmann, J.} \emph{et~al.}
\newblock \bibinfo{title}{The harvard clean energy project: large-scale
  computational screening and design of organic photovoltaics on the world
  community grid}.
\newblock \emph{\bibinfo{journal}{The Journal of Physical Chemistry Letters}}
  \textbf{\bibinfo{volume}{2}}, \bibinfo{pages}{2241--2251}
  (\bibinfo{year}{2011}).

\bibitem{gomez2016design}
\bibinfo{author}{G{\'o}mez-Bombarelli, R.} \emph{et~al.}
\newblock \bibinfo{title}{Design of efficient molecular organic light-emitting
  diodes by a high-throughput virtual screening and experimental approach}.
\newblock \emph{\bibinfo{journal}{Nature Materials}}
  \textbf{\bibinfo{volume}{15}}, \bibinfo{pages}{1120--1127}
  (\bibinfo{year}{2016}).

\end{thebibliography}

\section*{Acknowledgements}

The Pande Group is broadly supported by grants from the NIH (R01 GM062868 and U19 AI109662).  B.R. was supported by the Fannie and John Hertz Foundation. E.N.F. was supported by NIH training grant T32 GM08294.  We would like to acknowledge the Stanford Computing Resources for providing access to the Sherlock and Xstream GPU clusters. Thanks to Aarthi Ramsundar for assistance with diagrams.

\section*{Author contributions statement}
J.G and B.R. developed the atomic convolution algorithm.  E.N.F. and B.R. developed the grid featurizer algorithm.  J.G., B.R., and E.N.F. planned the experiments.  J.G. collected and analyzed the data from experiments. J.G., B.R., and E.N.F. interpreted the data and wrote the manuscript.  V.S.P. advised the work and edited the manuscript.  J.G, B.R, E.N.F, and V.S.P. all approved the final revision of the manuscript.

\section*{Supplementary Information}
\begin{figure}
    \centering
    \includegraphics[width=\textwidth]{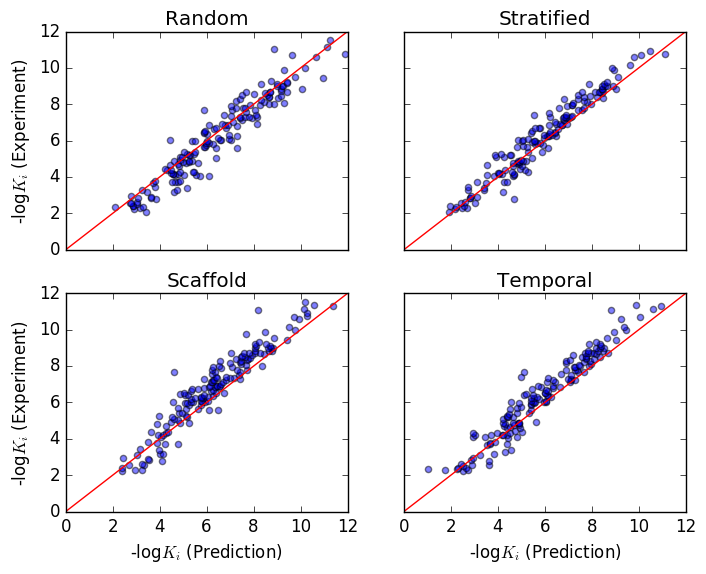}
    \caption{\textbf{ACNN Performance on core training set.} Comparison between experimental and predicted log$K_i$ evaluated on PDBBind core (a) Random (b) Stratified (c) Scaffold and (d) Temporal train sets.}
    \label{fig:acnn_core_train}
\end{figure}
\begin{figure}
    \centering
    \includegraphics[width=\textwidth]{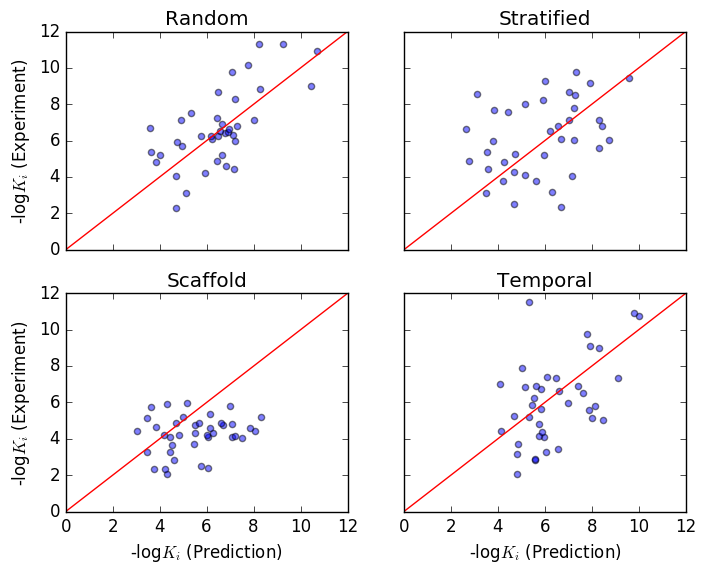}
    \caption{\textbf{ACNN Performance on core test set.} Comparison between experimental and predicted log$K_i$ evaluated on PDBBind core (a) Random (b) Stratified (c) Scaffold and (d) Temporal test sets.}
    \label{fig:acnn_core_test}
\end{figure}
\begin{figure}
    \centering
    \includegraphics[width=\textwidth]{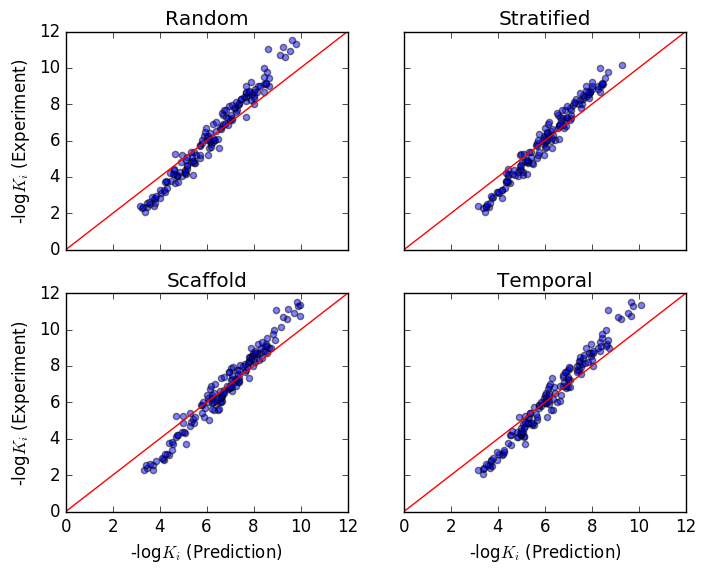}
    \caption{\textbf{GRID-RF Performance on core training set.} Comparison between experimental and predicted log$K_i$ evaluated on PDBBind core (a) Random (b) Stratified (c) Scaffold and (d) Temporal train sets.}
    \label{fig:acnn_core_train}
\end{figure}
\begin{figure}
    \centering
    \includegraphics[width=\textwidth]{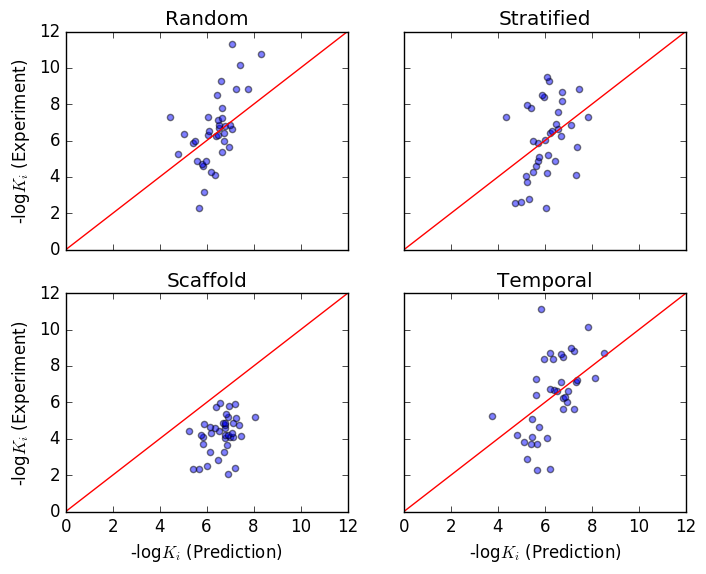}
    \caption{\textbf{GRID-RF Performance on core test set.} Comparison between experimental and predicted log$K_i$ evaluated on PDBBind core (a) Random (b) Stratified (c) Scaffold and (d) Temporal test sets.}
    \label{fig:acnn_core_test}
\end{figure}
\begin{figure}
    \centering
    \includegraphics[width=\textwidth]{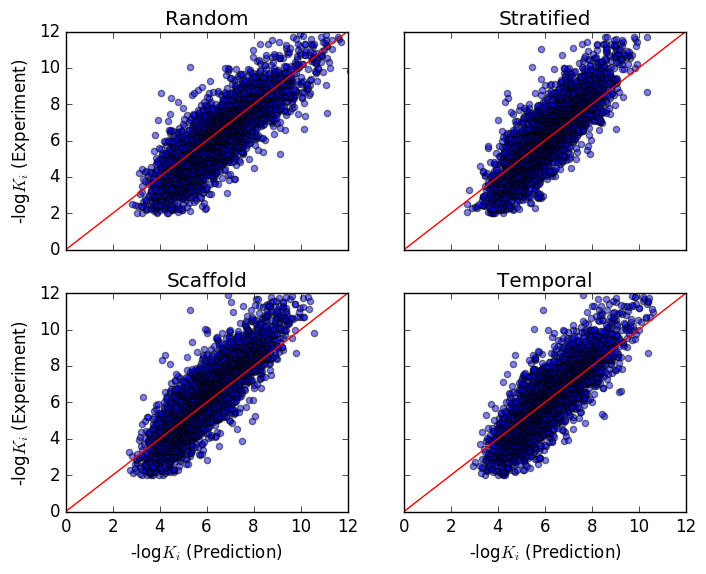}
    \caption{\textbf{ACNN Performance on refined training set.} Comparison between experimental and predicted log$K_i$ evaluated on PDBBind refined (a) Random (b) Stratified (c) Scaffold and (d) Temporal train sets.}
    \label{fig:acnn_core_train}
\end{figure}
\begin{figure}
    \centering
    \includegraphics[width=\textwidth]{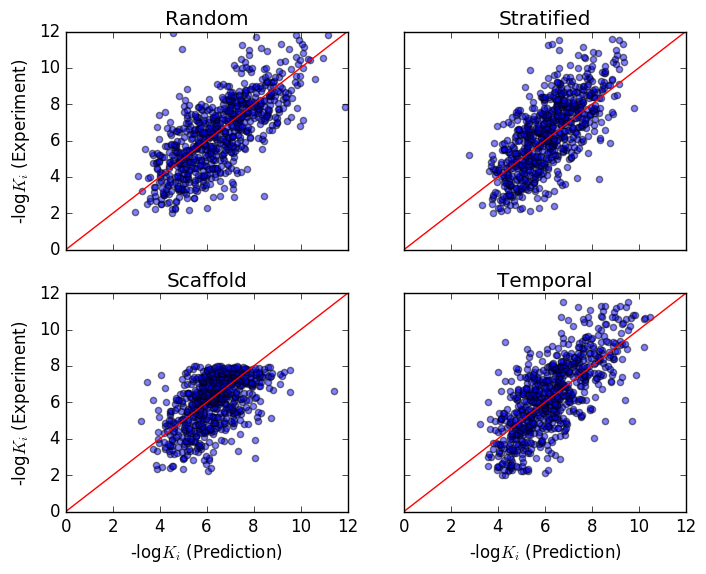}
    \caption{\textbf{ACNN Performance on refined test set.} Comparison between experimental and predicted log$K_i$ evaluated on PDBBind refined (a) Random (b) Stratified (c) Scaffold and (d) Temporal test sets.}
    \label{fig:acnn_core_test}
\end{figure}
\begin{figure}
    \centering
    \includegraphics[width=\textwidth]{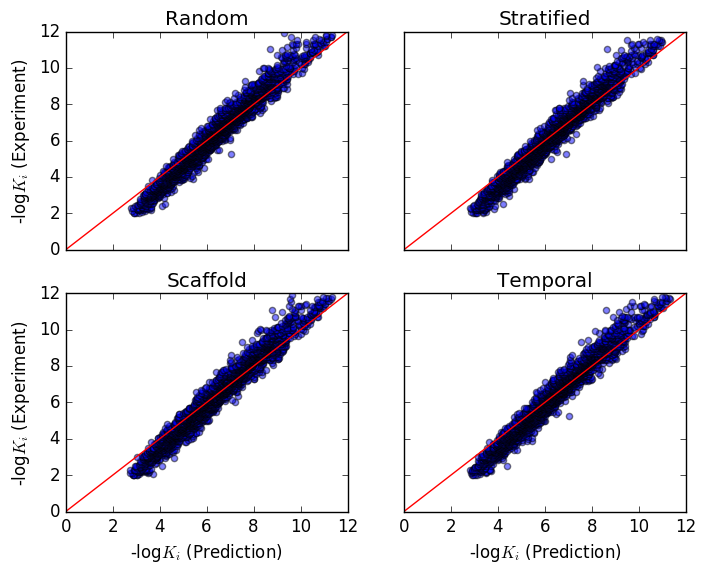}
    \caption{\textbf{GRID-RF Performance on refined training set.} Comparison between experimental and predicted log$K_i$ evaluated on PDBBind refined (a) Random (b) Stratified (c) Scaffold and (d) Temporal train sets.}
    \label{fig:acnn_core_train}
\end{figure}
\begin{figure}
    \centering
    \includegraphics[width=\textwidth]{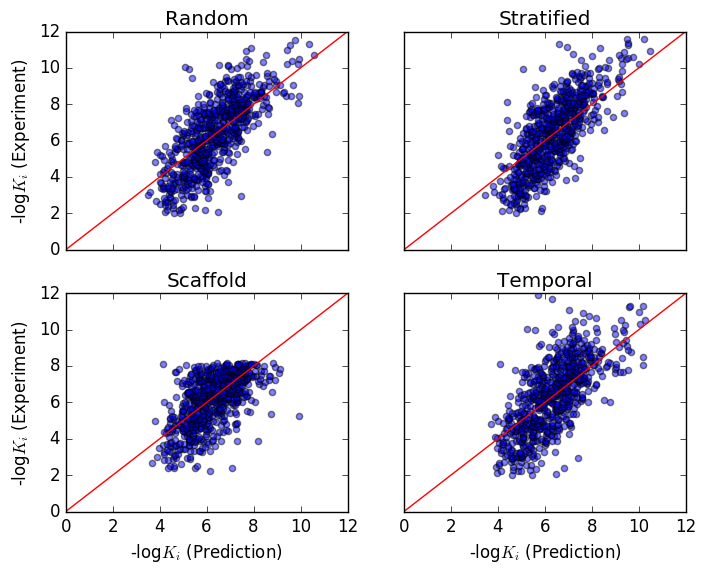}
    \caption{\textbf{GRID-RF Performance on refined test set.} Comparison between experimental and predicted log$K_i$ evaluated on PDBBind refined (a) Random (b) Stratified (c) Scaffold and (d) Temporal test sets.}
    \label{fig:acnn_core_test}
\end{figure}
\end{document}